\documentclass[11pt,a4paper]{article}
\usepackage[hyperref]{naaclhlt2019}
\usepackage{times}
\usepackage{latexsym}

\usepackage{helvet}  
\usepackage{courier}  
\usepackage{url}  
\usepackage{graphicx}  
\usepackage{amsfonts}
\usepackage{amsmath}
\usepackage{verbatim}
\usepackage{bm}
\usepackage{makecell}
\usepackage[font=small, labelfont=bf]{caption}
\usepackage[]{subcaption}
\usepackage{color, colortbl}
\usepackage{framed, enumitem}
\usepackage[colorinlistoftodos]{todonotes}
\newcommand{\ourmethod}{\textsc{Dolores}}

\definecolor{high}{HTML}{FFCE8E}

\aclfinalcopy 

\newcommand\BibTeX{B{\sc ib}\TeX}

\title{\ourmethod: \underline{D}eep C\underline{o}ntextua\underline{l}ized Kn\underline{o}wledge G\underline{r}aph \underline{E}mbedding\underline{s}}

 \author{Haoyu Wang\textsuperscript{1}, Vivek Kulkarni\textsuperscript{2}, William Yang Wang\textsuperscript{2}\\
 \textsuperscript{1}Department of Electronic Engineering, Shanghai Jiao Tong University\\
 \textsuperscript{2}Department of Computer Science, UC Santa Barbara\\
 why2011btv@sjtu.edu.cn, \{vvkulkarni,william\}@cs.ucsb.edu\\
 }

\date{}

\begin{document}
\maketitle
\begin{abstract}
We introduce a new method \ourmethod\ for learning knowledge graph embeddings that effectively captures contextual cues and dependencies among entities and relations. First, we note that short paths on knowledge graphs comprising of chains of entities and relations can encode valuable information regarding their contextual usage. We operationalize this notion by representing knowledge graphs not as a collection of triples but as a collection of entity-relation chains, and learn embeddings for entities and relations using deep neural models that capture such contextual usage. In particular, our model is based on Bi-Directional LSTMs and learn deep representations of entities and relations from constructed entity-relation chains. We show that these representations can very easily be incorporated into existing models to significantly advance the state of the art on several knowledge graph prediction tasks like link prediction, triple classification, and missing relation type prediction (in some cases by at least $\textbf{9.5\%}$).

\end{abstract}
\section{Introduction}

Knowledge graphs \cite{dong2014knowledge} enable structured access to world knowledge and form a key component of several applications like search engines, question answering systems and conversational assistants.  Knowledge graphs are typically interpreted as  comprising of discrete triples of the form \texttt{(entityA, relationX, entityB)} thus representing a relation (\texttt{relationX}) between \texttt{entityA} and \texttt{entityB}. However, one limitation of only a discrete representation of triples is that it does not easily enable one to infer similarities and potential relations among entities which may be missing in the knowledge graph. Consequently, one popular alternative is to learn dense continuous representations of entities and relations by embedding them in latent continuous vector spaces, while seeking to model the inherent structure of the knowledge graph. Most knowledge graph embedding methods can be classified into two major classes: one class which operates purely on triples like \textsc{RESCAL}~\cite{nickel2011three}, \textsc{TransE}~\cite{bordes2013translating}, \textsc{DistMult}~\cite{yang2014embedding}, \textsc{TransD}~\cite{ji2015knowledge}, \textsc{ComplEx}~\cite{trouillon2016complex}, \textsc{ConvE}~\cite{dettmers2017convolutional} and the second class which seeks to incorporate additional information (like multi-hops) \cite{wang2017knowledge}. Learning high-quality knowledge graph embeddings can be quite challenging given that (a) they need to effectively model the contextual usages of entities and relations (b) they would need to be useful for a variety of predictive tasks on knowledge graphs.

\begin{figure*}[htb!]
\centering
\begin{subfigure}{0.32\textwidth}
     \includegraphics[width=\textwidth]{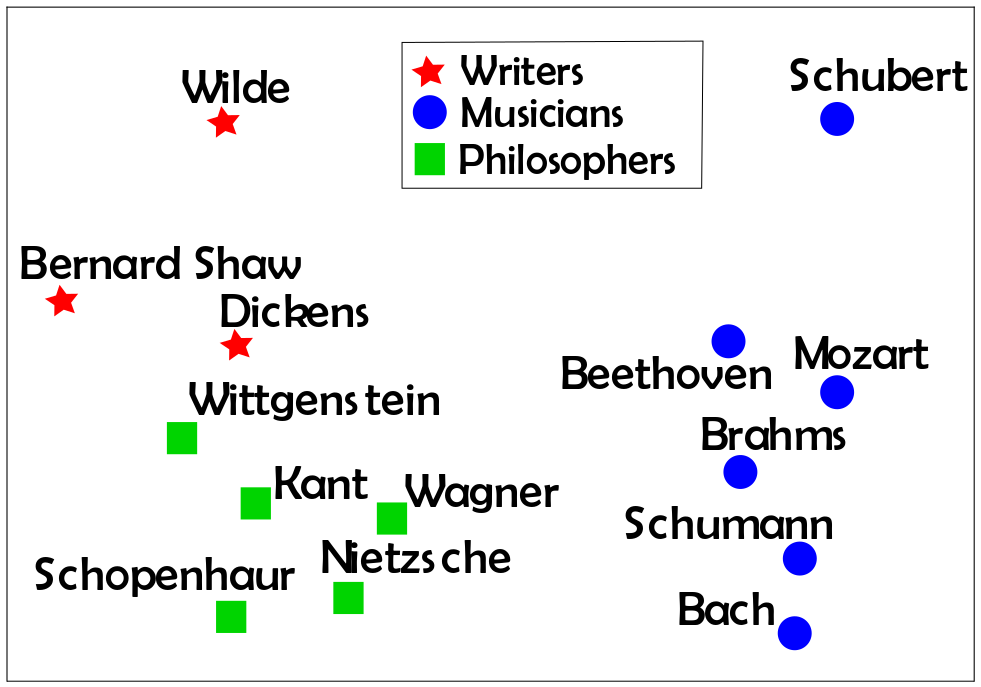}
    \caption{Context-independent}
    \label{fig:context_independent}
\end{subfigure} \hfill
\begin{subfigure}{0.32\textwidth}
\centering
    \includegraphics[width=\textwidth]{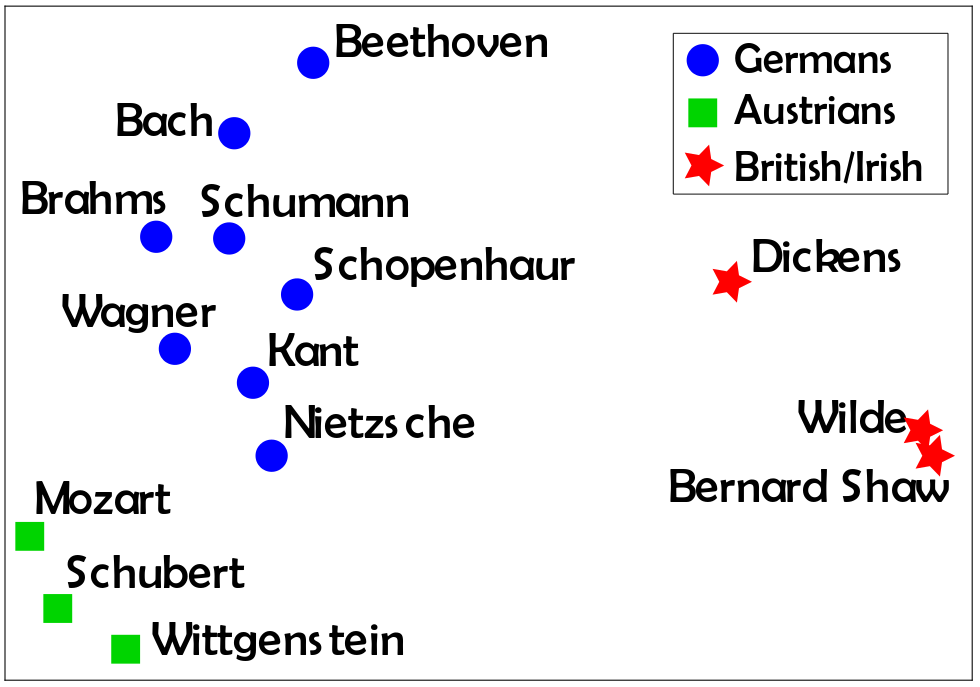}
    \caption{Nationality dependent}
    \label{fig:nationality}
\end{subfigure} \hfill
\begin{subfigure}{0.32\textwidth}
\centering
    \includegraphics[width=\textwidth]{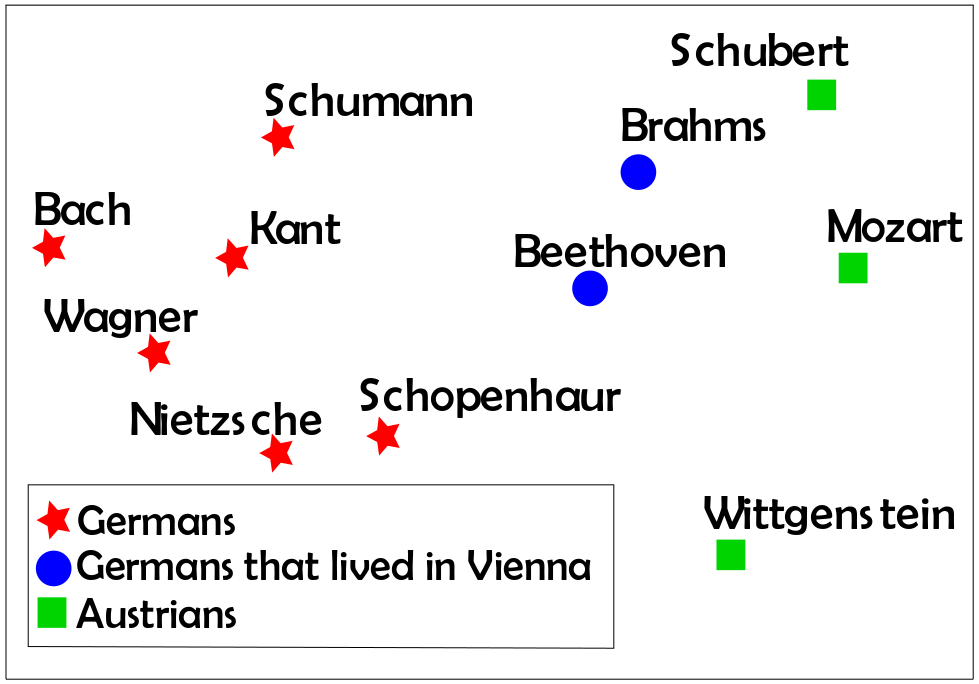}
    \caption{Place-lived dependent}
    \label{fig:place}
\end{subfigure}
\caption{Context independent and dependent embeddings learned by \ourmethod. (a) shows context-independent representations of writers (red), philosophers (green), and musicians (blue); (b) shows contextual embeddings with relation `/people/nationality': Austrians (green), Germans (blue), British/Irish (red). (c) shows contextual embeddings with relation `/people/place\_lived/location', we can see that Beethoven and Brahms (in blue), though Germans (other Germans are in red), lived in Vienna, Austria (Austrians are in green) and lie in between Germans and Austrians.}
\label{fig:crown_jewel}
\end{figure*}

In this paper, we present a new type of knowledge graph embeddings called \ourmethod\ that are both \emph{deep} and \emph{contextualized}. \ourmethod\ learns both context-independent and context-dependent embeddings of entities and relations through a deep neural sequential model. Figure \ref{fig:crown_jewel} illustrates the deep contextualized representations learned. Note that the contextually independent entity embeddings (see Figure \ref{fig:context_independent}) reveal three clusters of entities: writers, philosophers, and musicians. The contextual dependent embeddings in turn effectively account for specific relations. In particular, the context-dependent representations under the relation \texttt{nationality} now nicely cluster the above entities by nationality namely \texttt{Austrians, Germans, and British/Irish}. Similarly Figure \ref{fig:place} shows contextual embeddings given the relation \texttt{place-lived}. Note that these embeddings correctly capture that even though Beethoven and Brahms being Germans, they lived in Vienna and are closer to other Austrian musicians like Schubert.

Unlike most knowledge graph embeddings like \textsc{TransD, TransE} \cite{bordes2013translating,ji2015knowledge} etc. which are typically learned using shallow models, the representations learned by \ourmethod\ are deep: dependent on an entire path (rather than just a triple), are functions of internal states of a Bi-Directional LSTM and composed of representations learned at various layers potentially capturing varying degrees of abstractions. \ourmethod\ is inspired by recent advances in learning word representations (word embeddings) from deep neural language models using Bi-Directional LSTMs \cite{peters2018deep}.  In particular, we derive connections between the work of Peters et al. (\citeyear{peters2018deep}) who learn \emph{deep contextualized} word embeddings from sentences using a Bi-Directional LSTM based language model and \emph{random walks on knowledge graphs}. These connections enable us to propose new ``deep contextualized'' knowledge graph embeddings which we call \ourmethod\ embeddings.

Knowledge Embeddings learned using \ourmethod\ can easily be used as input representations for predictive models on knowledge graphs. More importantly, when existing predictive models use input representations for entities and relations, we can easily replace those representations with \ourmethod\ representations and significantly improve the performance of existing models. Specifically, we show that \ourmethod\ embeddings advance the state-of-the-art models on various tasks like link prediction, triple classification and missing relation type prediction.

To summarize, our contributions are as follows:
\begin{enumerate}
\item We present a new method \ourmethod\ of learning deep contextualized knowledge graph embeddings using a deep neural sequential model.
\item These embeddings are functions of hidden states of the deep neural model and can capture both \emph{context-independent} and \emph{context-dependent} cues. 
\item We show empirically that \ourmethod\ embeddings can easily be incorporated into existing predictive models on knowledge graphs to advance the state of the art on several tasks like link prediction, triple classification, and missing relation type prediction. 
\end{enumerate}

\section{Related Work}
Extensive work exists on knowledge graph embeddings dating back to Nickel, Tresp, and Kriegel (\citeyear{nickel2011three}) who first proposed \textsc{Rescal} based on a matrix factorization approach. Bordes et al. (\citeyear{bordes2013translating}) advanced this line of work by proposing the first translational model \textsc{TransE} which seeks to relate the head and tail entity embeddings by modeling the relation as a translational vector. This culminated in a long series of new knowledge graph embeddings all based on the translational principle with various refinements \cite{wang2014knowledge,lin2015learning,ji2015knowledge,yang2014embedding,trouillon2016complex,Nickel2017PoincarEF,minervini2017regularizing,xiao2017ssp,ma2017transt,chen2017learning,chen2018on2vec}. Some recently proposed models like \textsc{ManiFoldE} \cite{xiao2015one} attempt to learn knowledge graph embeddings as a manifold while embeddings like \textsc{HolE} \cite{nickel2011three} derive inspiration from associative memories. Furthermore, with the success of neural models, models based on convolutional neural networks have been proposed like \cite{dettmers2017convolutional,shi2017proje} to learn knowledge graph embeddings. Other models in this class of models include \textsc{ConvKB} \cite{nguyen2018novel} and \textsc{KBGAN} \cite{cai2018kbgan}. There has been some work on incorporating additional information like entity types, relation paths etc. to learn knowledge graph representations. Palumbo et al. (\citeyear{palumbo2018knowledge}) use \textsc{node2vec} to learn embeddings of entities and items in a knowledge graph. A notable class of methods called ``path-ranking'' based models directly model paths between entities as features. Examples include Path Ranking Algorithm (PRA) \cite{lao2012reading}, PTransE \cite{lin2015learning} and models based on recurrent neural networks \cite{neelakantan2015compositional,das2017chains}. Besides, Das et al. (\citeyear{das2017go}) propose a reinforcement learning method that addresses practical task of answering questions where the relation is known, but only one entity. Hartford et al. (\citeyear{hartford2018dmias}) model interactions across two or more sets of objects using a parameter-sharing scheme. 

While most of the above models except for the recurrent-neural net abased models above are shallow our model \ourmethod\ differs from all of these works and especially that of Palumbo et al. (\citeyear{palumbo2018knowledge}) in that we learn deep contextualized knowledge graph representations of entities and relations using a deep neural sequential model. The work that is closest to our work is that of Das et al. (\citeyear{das2017chains}) who directly use an RNN-based architecture to model paths to predict missing links. We distinguish our work from this in the following key ways: (a) First, unlike Das et al. (\citeyear{das2017chains}), our focus is not on path reasoning but on learning rich knowledge graph embeddings useful for a variety of predictive tasks. Moreover while Das et al. (\citeyear{das2017chains}) need to use paths generated from PRA that typically correlate with relations, our method has no such restriction and only uses paths generated by generic random walks greatly enhancing the scalability of our method. In fact, we incorporate \ourmethod\ embeddings to improve the performance of the model proposed by Das et al. (\citeyear{das2017chains}). (b) Second, and most importantly we learn knowledge graph embeddings at multiple layers each potentially capturing different levels of abstraction.  (c) Finally, while we are inspired by the work of Peters et al. (\citeyear{peters2018deep}) in learning deep word representations, we build on their ideas by drawing connections between knowledge graphs and language modeling \cite{peters2018deep}. In particular, we propose methods to use random walks over knowledge graphs in conjunction with the machinery of deep neural language modeling to learn powerful deep contextualized knowledge graph embeddings that improve the state of the art on various knowledge graph tasks. 

\section{Method and Models}
\subsection{Problem Formulation}
Given a knowledge graph $G=(E, R)$ where $E$ denotes the set of entities and $R$ denotes the set of relations among those entities, we seek to learn $d$-dimensional embeddings of the entities and relations. In contrast to previous knowledge graph embedding methods like \cite{bordes2013translating,wang2014knowledge,ji2015knowledge,lin2015learning,trouillon2016complex} which are based on shallow models and operates primarily on triples, our method \ourmethod\ uses a \emph{deep} neural model to learn \emph{``deep'' and ``contextualized''} knowledge graph embeddings.

Having formulated the problem, we now describe \ourmethod. \ourmethod\ consists of two main components:
\begin{enumerate}
\item \textbf{Path Generator} This component is responsible for generating a large set of entity-relation chains that reflect the varying contextual usages of entities and relations in the knowledge graph. 
\item \textbf{Learner} This component is a deep neural model that takes as input entity-relation chains and learns entity and relation embeddings which are weighted linear combination of internal states of the model thus capturing context dependence. 
\end{enumerate}

Both of the above components are motivated by recent advances in learning deep representations of words in language modeling. We motivate this below and also highlight key connections that enable us to build on these advances to learn \ourmethod\ knowledge graph embeddings.

\subsection{Preliminaries}
\paragraph{Language Modeling} Recall that the goal of a language model is to estimate the likelihood of a sequence of words: $w_{1}, w_{2}, \cdots,w_{n}$ where each word $w_{i}$ is from a finite vocabulary $\mathcal{V}$. Specifically, the goal of a \emph{forward language model} is to model $\text{Pr}(w_{i}|w_{1}, w_{2}, \cdots,w_{i-1})$. While, traditionally this has been modeled using count-based ``n-gram based" models \cite{manning1999foundations,jurafsky2000speech}, recently deep neural models like LSTMs and RNN's have been used to build such language models. As noted by Peters et al. (\citeyear{peters2018deep}), a forward language model implemented using an LSTM of ``L'' layers works as follows: At each position $k$, each LSTM layer outputs a \emph{context-dependent} representation denoted by $\overrightarrow{h_{k,j}}$ corresponding to the $j^{th}$ layer of the LSTM. The top-most layer of the LSTM is then fed as input to a softmax layer of size $|\mathcal{V}|$ to predict the next token. Similarly, a backward language model which models $\text{Pr}(w_{i}|w_{i+1}, w_{i+2}, \cdots,w_{n})$ can be implemented using a ``\emph{backward}'' LSTM  producing similar representations. A Bi-Directional LSTM just combines both forward and backward directions and seeks to jointly maximize the log-likelihood of the forward and backward directional language model objectives. 

We note that these context-dependent representations learned by the LSTM at each layer have been shown to be useful as ``deep contextual'' word representations in various predictive tasks in natural language processing \cite{peters2018deep}. In line with this trend, we will also use deep neural sequential models more specifically Bi-Directional LSTMs to learn \ourmethod\ embeddings. We do this by generalizing this approach to graphs by noting connections first noted by Perozzi, Al-Rfou, and Skiena (\citeyear{perozzi2014deepwalk}).

\paragraph{Connection between Random Walks on Graphs and Sentences in Language} Since the input to a language model is a large corpus or set of sentences, one can generalize language modeling approaches to graphs by noting that the analog of a sentence in graphs is a ``random walk''. More specifically, note that a truncated random walk of length T starting from a node ``v'' is analogous to a sentence and effectively captures the context of ``v" in the network. More precisely, the same machinery used to learn representations of words in language models can now be adapted to learn deep contextualized representations of knowledge graphs.

This can easily be adapted to knowledge graphs by constructing paths of entities and relations. In particular, a random walk on a knowledge graph starting at entity $e_{1}$ and ending at entity $e_{k}$ is a sequence of the form $e_{1}, r_{1}, e_{2}, r_{2}, \cdots,e_{k}$ representing the entities and the corresponding relations between $e_{1}$ and $e_{k}$ in the knowledge graph. Moving forward we denote such a path of entities and relations by $\mathbf{q}=(e_{1},r_{1}, e_{2}, r_{2}, \cdots,e_{k})$. We generate a large set of such paths from the knowledge graph G by performing several random walks on it which in turn yields a corpus of ``sentences'' \textsl{S} needed for ``language modeling''.

\subsection{\ourmethod: Path Generator}
Having motivated the model and discussed preliminaries, we now describe the first component of \ourmethod\ -- the path generator.

Let \textsl{S} denote the set of entity-relation chains obtained by doing random walks in the knowledge graph. We adopt a component of \textsc{node2vec} \cite{grover2016node2vec} to construct \textsl{S}. In particular, we perform a $2^{nd}$ order random walk with two parameters \textit{p} and \textit{q} that determine the degree of breadth-first sampling and depth-first sampling. Specifically as Grover and Leskovec (\citeyear{grover2016node2vec}) described, \textit{p} controls the likelihood of immediately revisiting a node in the walk whereas \textit{q} controls whether the walk is biased towards nodes close to starting node or away from starting node. We emphasize that while \textsc{node2vec} has additional steps to learn dense continuous representations of nodes, we adopt only its first component to generate a corpus of random walks representing paths in knowledge graphs. 

\subsection{\ourmethod: Learner}
\begin{figure*}[htb!]
\centering
\includegraphics[width=0.86\textwidth]{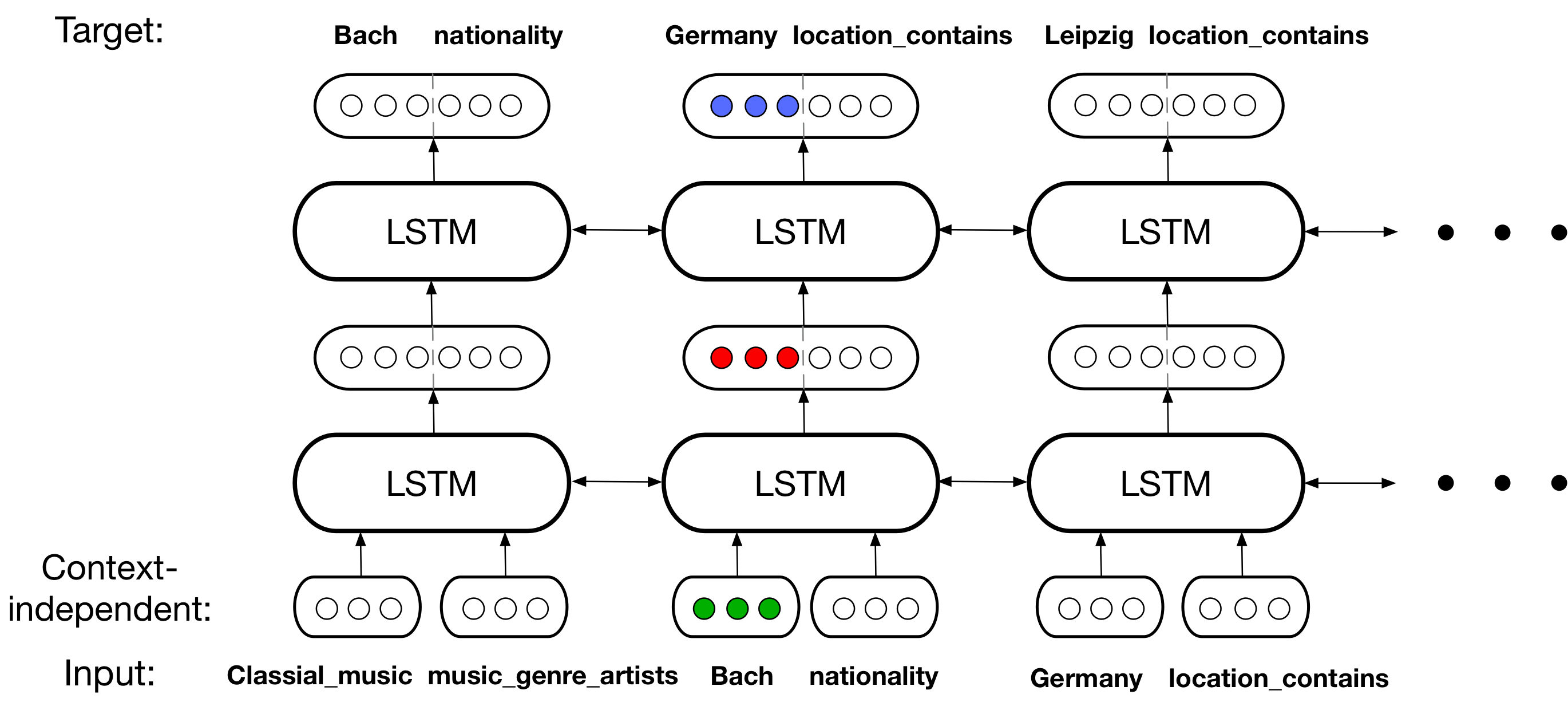}
\caption{Unrolled RNN architecture. The input to the deep Bi-Directional LSTM is entity-relation chains generated from random walks in KG. At each time step, the LSTM consumes the concatenation of entity and relation vectors. The ``target" shown in the figure is the training target for the forward direction. Learned contextual representation of entity ``\textbf{Bach}" is the concatenation of the embedding layer (green) and a linear combination of internal states of deeper layers (red and blue).}
\label{fig:unrolled}
\end{figure*}
Having generated a set of paths on knowledge graphs representing local contexts of entities and relations, we are now ready to utilize the machinery of language modeling using deep neural networks to learn \ourmethod\ embeddings.

While traditional language models model a sentence as a sequence of words, we adopt the same machinery to model knowledge graph embeddings as follows: (a) A word is an (entity, relation) tuple, (b) we model a sentence as a path consisting of (entity, relation) tuples.
Note that we have already established how to generate such paths from the knowledge graph using the path generator component.

Given such paths, we would like to model the probability of an entity-relation pair given the history and future context by a Bi-Directional Long Short-Term Memory network. In particular, the forward direction LSTM models:
\begin{equation*}
\text{Pr}([e_1,r_1], [e_2,r_2], \cdots, [e_N, r_N]) =  
\end{equation*}
\begin{equation}
\prod_{t=1}^{N} \text{Pr}( [e_t, r_t] \mid [e_1, r_1], [e_2, r_2], \cdots, [e_{t-1}, r_{t-1}] ).
\end{equation}
Similarly, the backward direction LSTM models:
\begin{equation*}
\text{Pr}([e_1,r_1], [e_2,r_2], \cdots, [e_N, r_N]) = 
\end{equation*}
\begin{equation}
\prod_{t=1}^{N} \text{Pr}( [e_t, r_t] \mid [e_{t+1}, r_{t+1}], \cdots, [e_{N}, r_{N}] ).
\end{equation}
Figure \ref{fig:unrolled} illustrates this succinctly. At each time-step \textit{t}, we deal with an entity-relation pair [$e_t$, $r_t$]. We first map one-hot vectors of the $e_t$ and $r_t$ to an embedding layer, concatenate them to obtain context-independent representations which are then passed through \textit{L} layers of a Bi-Directional LSTM. Each layer of LSTM outputs the pair's context-dependent representation $\overrightarrow{h_{t,i}}$, $\overleftarrow{h_{t,i}}$, where i=1, 2, $\cdots$, \textit{L}. Finally, the output of the top layer of LSTM, $\overrightarrow{h_{t,L}}$, $\overleftarrow{h_{t,L}}$, is used to predict the next pair [$e_{t+1}$, $r_{t+1}$] and [$e_{t-1}$, $r_{t-1}$] respectively using a softmax layer.  Formally, we jointly maximize the log likelihood of the forward and backward directions:
\begin{equation}
\label{eq:objective}
\begin{split}
\sum_{t=1}^{N}\log\text{Pr}([e_t, r_t]\mid[e_1, r_1],\cdots,[e_{t-1}, r_{t-1}];\mathbf{\Theta_{F}})+\\
\sum_{t=1}^{N}\log\text{Pr}([e_t, r_t]\mid[e_{t+1}, r_{t+1}],\cdots,[e_{N}, r_{N}];\mathbf{\Theta_{B}}),
\end{split}
\end{equation}
where $\mathbf{\Theta_{F}}$=$(\theta_x, \overrightarrow{\theta_{LSTM}}, \theta_s)$ corresponds to the parameters of the embedding layer, forward-direction LSTM and the softmax layer respectively. Similarly $\mathbf{\Theta_{B}}$=$(\theta_x, \overleftarrow{\theta_{LSTM}}, \theta_s)$ corresponds to the similar set of parameters for the backward direction. Specifically, note that we share the parameters for the embedding and softmax layer across both directions. We maximize Equation \ref{eq:objective} by training the Bi-directional LSTMs using back-propagation through time.   

\paragraph{Extracting \ourmethod\ embeddings from the learner}
\begin{table*}[htb!]
\centering
\begin{tabular}{l|cc|cc}
\hline
\textsc{Task}                       & \multicolumn{2}{c|}{\textsc{Previous SOTA}} & \begin{tabular}[c]{@{}c@{}}\ourmethod + \\ \textsc{Baseline}\end{tabular} & \begin{tabular}[c]{@{}c@{}}\textsc{Increase} \\ (\textsc{Absolute}/\\ \textsc{Relative})\end{tabular} \\ \hline
Link Prediction (head) & \cite{nguyen2018novel}    & 35.5     & 37.5                                                          & 2.0 / 3.1\%                                                                \\
Link Prediction (tail) & \cite{nguyen2018novel}     & 44.3     & 48.7                                                          & 4.4 / 7.9\%                                                                \\
Triple Classification      & \cite{nguyen2018novel}     & 88.20    & 88.40                                                         & 0.20 / 1.7\%                                                               \\
Missing Relation Type    & \cite{das2017chains}      & 71.74    & 74.42                                                         & 2.68 / 9.5\%                                                               \\ \hline
\end{tabular}
\caption{Summary of results of incorporating \ourmethod\ embeddings on state-of-the-art models for various tasks. Note that in each case, simply incorporated \ourmethod\ results in a significant improvement over the state of the art in various tasks like link prediction, triple classification and missing relation type prediction sometimes by as much as $\textbf{9.5\%}$.}
\label{tab:summary}
\end{table*}
\begin{table*}[htb!]
\renewcommand\arraystretch{1.1}
\centering
\resizebox{2.1\columnwidth}{!}{%
\begin{tabular}{|l|ccccccccc|}
\hline
                        & \multicolumn{9}{c|}{\textsc{FB15K237}}                                                                                       \\ \cline{2-10} 
                        & \multicolumn{3}{c|}{\textsc{head}}                  & \multicolumn{3}{c|}{\textsc{tail}}                   & \multicolumn{3}{c|}{\textsc{Avg.}} \\ \cline{2-10} 
\textsc{Method}                  & \textsc{MRR}   & \textsc{MR}  & \multicolumn{1}{c|}{\textsc{HITS@10}} & \textsc{MRR}   & \textsc{MR}  & \multicolumn{1}{c|}{\textsc{HITS@10}} & \textsc{MRR}     & \textsc{MR}   & \textsc{HITS@10}  \\ \hline
\textsc{TransE}                  & 0.154 & 651 & \multicolumn{1}{c|}{0.294}   & 0.332 & 391 & \multicolumn{1}{c|}{0.524}   & 0.243   & 521  & 0.409    \\ 
\textsc{ConvE}                      & 0.204 & 375 & \multicolumn{1}{c|}{0.366}   & 0.408 & 189 & \multicolumn{1}{c|}{0.594}& 0.306   & 283  & 0.480    \\ 
\textsc{ConvKB} (\textbf{SOTA})                  & 0.355 & 348 & \multicolumn{1}{c|}{0.459}   & 0.443 & 178 & \multicolumn{1}{c|}{0.572}   & 0.399   & 263  & 0.515    \\ \hline 
\textsc{ConvKB} (+ \textbf{\ourmethod}) & \textbf{0.375} & \textbf{316} & \multicolumn{1}{c|}{\textbf{0.476}}   & \textbf{0.487} & \textbf{158} & \multicolumn{1}{c|}{\textbf{0.596}}   & \textbf{0.431}   & \textbf{237}  & \textbf{0.536}    \\ \hline
\rowcolor{high}
\textsc{Improvement (relative \%)} & 3.10\% & 9.20\% & \multicolumn{1}{c|}{3.14\%} & 7.90\% & 11.23\% & \multicolumn{1}{c|}{5.61\%} & 5.32\% & 9.89\% &4.33\%  \\ \hline
\end{tabular}
}
\caption{Performance of incorporating \ourmethod\ on state-of-the-art model for link prediction. Note that we consistently and significantly improve the current state of the art in both subtasks: head entity and tail entity prediction (in some cases by at least $\textbf{9\%}$). \emph{For all metrics except MR (mean rank) higher is better.}}
\label{tab:link_pred}
\end{table*}
After having estimated the parameters of the \ourmethod\ learner, we now extract the context-independent and context-dependent representations for each entity and relation and combine them to obtain \ourmethod\ embeddings. 
More specifically, \ourmethod\ embeddings are task specific combination of the context-dependent and context-independent representations learned by our learner.  Note that our learner (which is an $L$-layer Bi-Directional LSTM) computes a set of $2L + 1$ representations for each entity-relation pair which we denote by:
\begin{equation*}
R_t = [ x_t, \overrightarrow{h_{t,i}}, \overleftarrow{h_{t,i}} \mid i = 1, 2, \cdots, \textit{L} ],
\end{equation*}
where $x_t$ is the context-independent embedding and $\overrightarrow{h_{t,i}}, \overleftarrow{h_{t,i}}$ correspond to the context-dependent embeddings from layer $i$.

Given a downstream model, \ourmethod\ learns a weighted linear combination of the components of $R_t$ to yield a single vector for use in the embedding layer of the downstream model. In particular
\begin{equation}
\texttt{\ourmethod}_t = [ x_t , \sum_{i=1}^{L} \lambda_i h_{t,i} ],
\end{equation}
where we denote 
$h_{t,i}$ =  [ $\overrightarrow{h_{t,i}}$, $\overleftarrow{h_{t,i}}$ ] and $\lambda_{i}$ denote task specific learnable weights of the linear combination.

\paragraph{Incorporating \ourmethod\ embeddings into existing predictive models on Knowledge Graphs} While it is obvious that our embeddings can be used as features for new predictive models, it is also very easy to incorporate our learned \ourmethod\ embeddings into existing predictive models on knowledge graphs. The only requirement is that the model accepts as input, an embedding layer (for entities and relations). If a model fulfills this requirement (which a large number of neural models on knowledge graphs do), we can just use \ourmethod\ embeddings as a drop-in replacement. We just initialize the corresponding embedding layer with \ourmethod\ embeddings. In our evaluation below, we show how to improve several state-of-the-art models on various tasks simply by incorporating \ourmethod\ as a drop-in replacement to the original embedding layer.

\section{Experiments}

We evaluate \ourmethod\ on a set of $4$ different prediction tasks on knowledge graphs. In each case, simply adding \ourmethod\ to existing state-of-the-art models improves the state of the art performance significantly (in some cases by at least $9.5\%$) which we show in Table \ref{tab:summary}. While we primarily show that we can advance the state-of-the-art model by incorporating \ourmethod\ embeddings as a ``drop-in'' replacement, for the sake of completeness, we also report raw numbers of other strong baseline methods (like \textsc{TransD, TransE, KBGAN, ConvE}, and \textsc{ConvKB}) to place the results in context.
We emphasize that our method is very generic and can be used to improve the performance of a large class of knowledge graph prediction models. In the remainder of the section, we briefly provide high-level overviews of each task and summarize results for all tasks considered. 

\subsection{Experimental Settings for \ourmethod}
Here, we outline our model settings for learning \ourmethod\ embeddings. We generate 20 chains for each node in the knowledge graph, with the length of each chain being 21 (10 relations and $11$ entities appear alternately)\footnote{The total number of chains generated for the training, development, and test sets is at most $300K$. Also, we observed no significant differences with larger chains.}. Our model uses \textit{L} = 4  LSTM layers with $512$ units and $32$ dimension projections (projected values are clipped element-wise to within $[-3, 3]$).  We use residual connections between layers and the batch size is set to $1024$ during the training process. We train \ourmethod\ for 200 epochs on corresponding datasets with dropout (with the dropout probability is set $0.1$). Finally, we use Adam as the optimizer with appropriately chosen learning rates based on a validation set. 

\subsection{Evaluation Tasks}

We consider three tasks, link prediction, triple classification, and predicting missing relation types \cite{das2017chains}:

\begin{itemize}
\item \textbf{Link Prediction} A common task for knowledge graph completion is link prediction, aiming to predict the missing entity when the other two parts of a triplet (\textit{h, r, t}) are given. In other words, we need to predict \textit{t} given (\textit{h, r}) -- \emph{tail-entity prediction} or predict \textit{h} given (\textit{r, t}) -- \emph{head entity prediction}. In-line with prior work \cite{dettmers2017convolutional}, we report results on link prediction in terms of Mean Reciprocal Rank (MRR), Mean Rank (MR) and Hits@10 on the FB15K-237 dataset in the filtered setting on both sub-tasks: (a) head entity prediction and (b) tail entity prediction.  Our results are shown in Table \ref{tab:link_pred}. Note that the present state-of-the-art model, \textsc{ConvKB} achieves an MRR of $(0.375, 0.487)$ on the head and tail link prediction tasks. Observe that simply incorporating \ourmethod\ significantly improves the head and tail entity prediction performance by $3.10\%$ and $7.90\%$ respectively.  Similar improvements are also observed on other metrics like \textsc{Mean Rank} (MR: lower is better) and \textsc{Hits@10} (higher is better).  

\item \textbf{Triple Classification} The task of triple classification is to predict whether a triple (\textit{h, r, t}) is correct or not. Triple classification is a binary classification task widely explored by previous work \cite{bordes2013translating,wang2014knowledge,lin2015learning}. Since evaluation of classification needs negative triples, we choose WN11 and FB13, two benchmark datasets and report the results of our evaluation in Table \ref{tab:triples}. We note that the present state of the art is the \textsc{ConvKB} model. When we add \ourmethod\ to the  \textsc{ConvKB} model with our embeddings, observe that we improve the average performance of the state-of-the-art model \textsc{ConvKB} slightly by $0.20$ points (from $88.20$ to $88.40$). We believe the improvement achieved by adding \ourmethod\ is smaller in terms of absolute size because the state-of-the-art model already has excellent performance on  this task ($88.20$) suggesting a much slower improvement curve. 
\begin{table}[ht]
\begin{tabular}{l|ll|l}
\hline
\textsc{Method}                 & \textsc{WN11} & \textsc{FB13} & \textsc{Avg.} \\ \hline
\textsc{TransE}                 & 86.5 & 87.5 & 87.00 \\ 
\textsc{TransR}                 & 85.9 & 82.5 & 84.20 \\ 
\textsc{TransD}                 & 86.4 & \underline{89.1} & 87.75 \\ 
\textsc{TransG}                 & 87.4 & 87.3 & 87.35 \\ 
\textsc{ConvKB} (\textbf{SOTA})  & \textbf{87.6} & 88.8 & \underline{88.20} \\ 
\textsc{ConvKB} (+ \textbf{\ourmethod}) & \underline{87.5} & \textbf{89.3} & \textbf{88.40} \\ \hline
\end{tabular}
\caption {Experimental results of triple classification on WN11 and FB13 test sets. TransE is implemented by Nguyen et al. (\citeyear{nguyenconvolutional}). Except for \textsc{ConvKB}(+ \ourmethod), other results are from Nguyen et al. (\citeyear{nguyenconvolutional}). Bold results are the best ones and underlined results are second-best.}
\label{tab:triples}
\end{table}
\item \textbf{Missing Relation Types}
The goal of the third task is to reason on the paths connecting an entity pair to predict missing relation types. We follow Das et al. (\citeyear{das2017chains}) and use the same dataset released by Neelakantan, Roth, and McCallum (\citeyear{neelakantan2015compositional}) which is a subset of \textsc{FreeBase} enriched with information from \textsc{ClueWeb}. The dataset consists of a set of triples (\textit{$e_1$, r, $e_2$}) and the set of paths connecting the entity pair (\textit{$e_1$, $e_2$}) in the knowledge graph. These triples are collected from \textsc{ClueWeb} by considering sentences that contain the entity pair in \textsc{Freebase}. Neelakantan et al. (\citeyear{neelakantan2015compositional}) infer the relation type by examining the phrase between two entities. We use the same evaluation criterion as used by Das et al. (\citeyear{das2017chains}) and report our results in Table \ref{tab:relation_types}. Note that the current state-of-the-art model from Das et al. (\citeyear{das2017chains}) yields a score of $71.74$. Adding \ourmethod\ to the model improves the score to $74.42$ yielding a $\textbf{9.5\%}$ improvement.
\end{itemize}
\begin{table}[ht]
\begin{tabular}{l|l}
\hline
\textsc{Model}  & \textsc{MAP} \\ \hline
PRA \cite{lao2011random}                       & 64.43 \\
PRA + Bigram (Neelakantan et.al \citeyear{neelakantan2015compositional})              & 64.93 \\
RNN-Path (Das et.al \citeyear{das2017chains}) & 68.43 \\
RNN-Path-entity (Das et.al \citeyear{das2017chains}) \textbf{SOTA} & 71.74 \\ 
\hline
RNN-Path-entity (+\textbf{\ourmethod}) & \textbf{74.42} \\ \hline
\end{tabular}
\caption{Results of missing relation type prediction. RNN-Path-entity \cite{das2017chains} is the \textbf{state of the art} which yields an improvement of $9.5\%$  (71.74 vs 74.42) on mean average precision (MAP) when incorporated with \ourmethod.}
\label{tab:relation_types}
\end{table}
\begin{figure*}[htb!]
\centering
\begin{subfigure}{0.45\textwidth}
     \includegraphics[width=\textwidth]{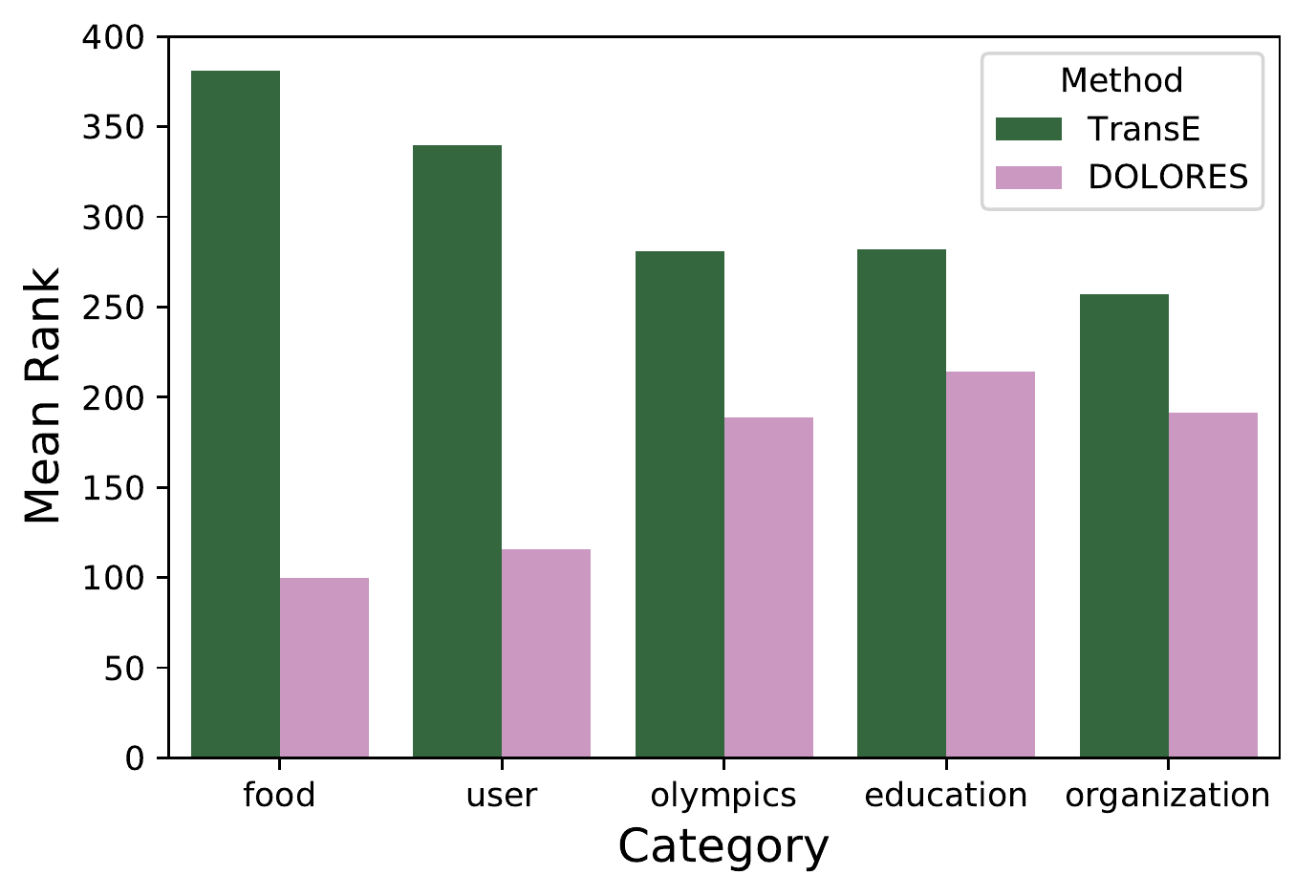}
    \caption{Best Prediction Categories for \ourmethod}
    \label{fig:best}
\end{subfigure}\hspace{2em}
\begin{subfigure}{0.45\textwidth}
    \includegraphics[width=\textwidth]{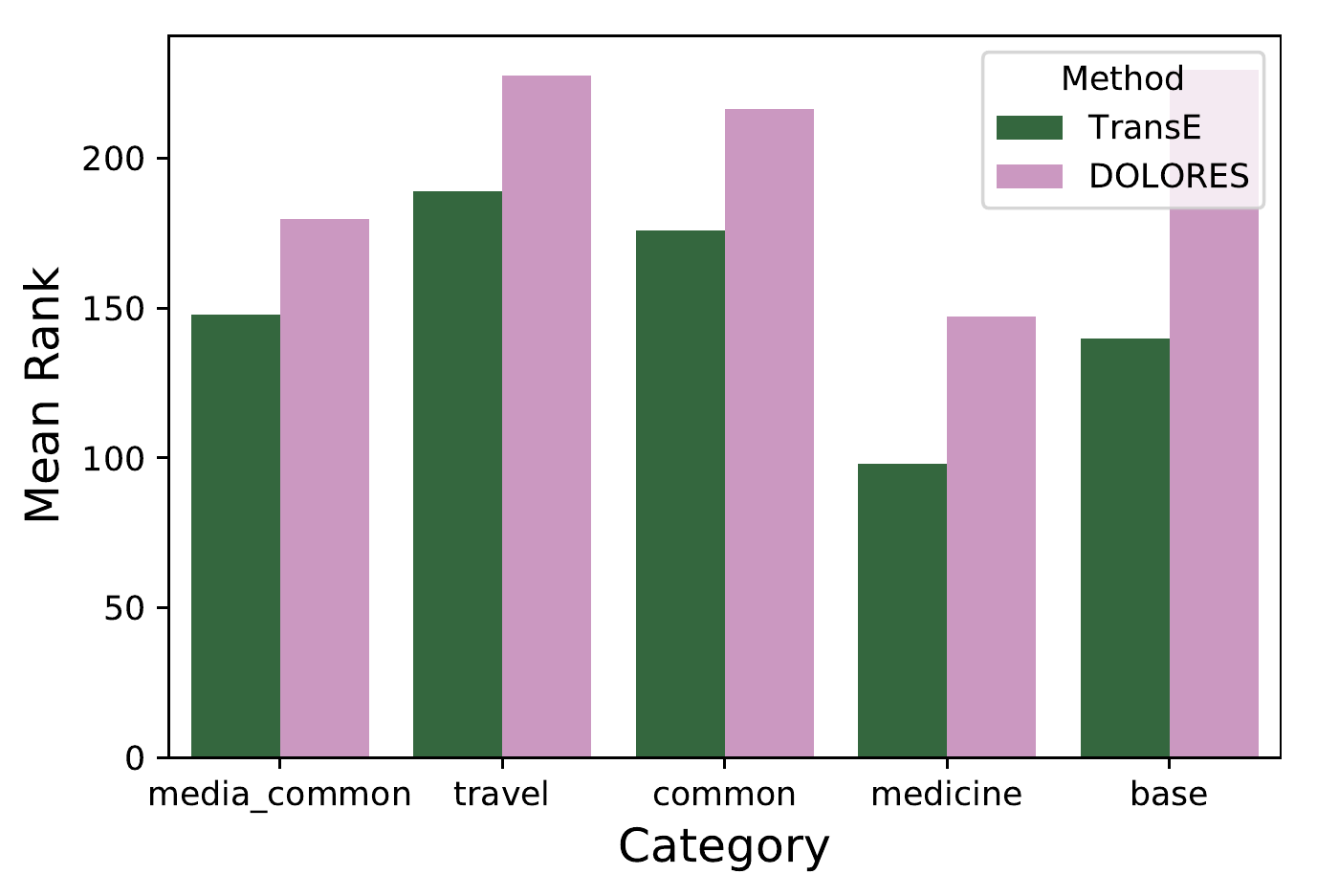}
    \caption{Worst Prediction Categories for \ourmethod
    }
    \label{fig:worst}
\end{subfigure}
\caption{Relation Categories with best and worst performance for \ourmethod\ in terms of mean rank (\textbf{lower is better}). \ourmethod\ performs exceedingly well on instances where the head entity is \textbf{specific} and tends to perform sub-optimally when the head entity is very \textbf{generic and broad} belonging to categories \texttt{base,common}. Please refer to Error Analysis section for detailed explanation and discussion.}
\label{fig:analysis}
\end{figure*}

Altogether, viewing the results on various tasks holistically, we conclude that simply incorporating \ourmethod\ into existing state-of-the-art models improves their performance and advances the state of the art on each of these tasks and suggests that our embeddings can be effective in yielding performance gains on a variety of predictive tasks.

\subsection{Error Analysis}
\label{sec:error}
In this section, we conduct an analysis of the predictions made when using \ourmethod\ on the link prediction tasks to gain insights into our learned embeddings. To do so, we group the test triples by the first component of their relation (the most abstract concept or level) and compute the mean rank of the tail entity output over each group. We compare this metric against what a simple baseline like \textsc{TransE} obtains. This enables us to identify overall statistical patterns that distinguish \ourmethod\ embeddings from baseline embeddings like \textsc{TransE}, which in turn boosts the performance of link prediction. 

Figure \ref{fig:analysis} shows the relation categories for which \ourmethod\ performed the best and the worst relative to baseline \textsc{TransE} embeddings in terms of mean rank (\textbf{lower is better}). In particular, Figure \ref{fig:best} shows the categories where \ourmethod\ performs the best and reveals categories like \texttt{food, user, olympics, organization}. Note for example, that for triples belonging under the \texttt{food} relation, \ourmethod\ outperforms baseline by \textsc{TransE} by a factor of $4$ in terms of mean rank. To illustrate this point, we show a few instances of such triples below:

\begin{itemize}
  \item \texttt{(A serious man, film-release-region, Hong Kong)}
  \item \texttt{(Cabaret, film-genre, Film Adaptation)}
  \item \texttt{(Lou Costello, people-profession, Comedian)}
\end{itemize}

Note that when the head entity is a very specific entity like \texttt{Louis Costello}, our method is very accurate at predicting the correct tail entity (in this case \texttt{Comedian}). \textsc{TransE}, on the other hand, makes very poor predictions on such cases. We believe that our method is able to model such cases better because our embeddings, especially for such specific entities, have captured the rich context associated with them from entire paths.  

In contrast, Figure \ref{fig:worst} shows the relation categories that \ourmethod\ performs the worst relative to \textsc{TransE}. We note that these correspond to very broad relation categories like \texttt{common,base,media\_common} etc. We list a few triples below to illustrate this point:
\begin{itemize}
  \item \texttt{(Psychology, \texttt{students majoring}, Yanni)}
  \item \texttt{(Decca Records, \texttt{music-record-label-art\\ist}, Jesseye Norman)}
  \item \texttt{(Priority Records, music-record-label-\\artist, Carole King)}
\end{itemize}
Note that such instances are indeed very difficult. For instance, given a very generic  entity like \texttt{Psychology} it is very difficult to guess that \texttt{Yanni} would be the expected tail entity. Altogether, our method is able to better model triples where the head entity is more specific compared to head entities which are very broad and general.

\section{Conclusion}
In this paper, we introduce \ourmethod, a new paradigm of learning knowledge graph embeddings where we learn not only contextual independent embeddings of entities and relations but also multiple context-dependent embeddings each capturing a different layer of abstraction. We demonstrate that by leveraging connections between three seemingly distinct fields namely: (a) large-scale network analysis, (b) natural language processing and (c) knowledge graphs we are able to learn rich knowledge graph embeddings that are deep and contextualized in contrast to prior models that are typically shallow. Moreover, our learned embeddings can be easily incorporated into existing knowledge graph prediction models to improve their performance. Specifically, we show that our embeddings are not only a ``drop-in'' replacement for existing models that use embedding layers but also significantly improve the state-of-the-art models on a variety of tasks, sometimes by almost $9.5\%$. Furthermore, our method is inherently online and scales to large data-sets. 

Our work also naturally suggests new directions of investigation and research into knowledge graph embeddings. One avenue of research is to investigate the utility of each layer's entity and relation embeddings in specific learning tasks.
As was noted by research in computer vision, deep representations learned on one dataset are effective and very useful in transfer-learning tasks \cite{huang2017transfer}.
A natural line of investigation thus revolves around precisely quantifying the effectiveness of these learned embeddings across models and data-sets.
Lastly, it would be interesting to see if such knowledge graph embeddings can be used in conjunction with natural language processing models used for relation extraction and named entity recognition from raw textual data. 

 Finally, in a broader perspective, our work introduces two new paradigms of modeling knowledge graphs. First, rather than view a knowledge graph as a collection of triples, we view it as a set of paths between entities. These paths can be represented as a collection of truncated random walks over the knowledge graph and encode rich contextual information between entities and relations. Second, departing from the hitherto well-established paradigm of mapping entities and relations to vectors in $\mathcal{R}^{d}$ via a mapping function, we learn multiple representations for entities and relations determined by the number of layers in a deep neural network. This enables us to learn knowledge graph embeddings that capture different layers of abstraction -- both context-independent and context-dependent allowing for the development of very powerful prediction models to yield superior performance on a variety of prediction tasks.
\bibliography{naaclhlt2019}
\bibliographystyle{acl_natbib}

\end{document}